\documentclass[letterpaper]{article} % DO NOT CHANGE THIS
\usepackage{aaai25}  % DO NOT CHANGE THIS
\usepackage{times}  % DO NOT CHANGE THIS
\usepackage{helvet}  % DO NOT CHANGE THIS
\usepackage{courier}  % DO NOT CHANGE THIS
\usepackage[hyphens]{url}  % DO NOT CHANGE THIS
\usepackage{graphicx} % DO NOT CHANGE THIS

\usepackage{amsmath}
\usepackage{amssymb}
\usepackage{color}
\usepackage{booktabs}
\usepackage{multirow}
\usepackage{pifont}

\usepackage{bbding}

\usepackage{natbib}  % DO NOT CHANGE THIS AND DO NOT ADD ANY OPTIONS TO IT
\usepackage{caption} % DO NOT CHANGE THIS AND DO NOT ADD ANY OPTIONS TO IT
\usepackage{algorithm}
\usepackage{algorithmic}

\usepackage{newfloat}
\usepackage{listings}
\DeclareCaptionStyle{ruled}{labelfont=normalfont,labelsep=colon,strut=off} % DO NOT CHANGE THIS
\floatstyle{ruled}
\newfloat{listing}{tb}{lst}{}
\floatname{listing}{Listing}
\title{Cross Fusion RGB-T Tracking with Bi-directional Adapter}
\author{
    Zhirong Zeng,
    Xiaotao Liu\thanks{Xiaotao Liu is the corresponding author.},
    Meng Sun,
    Hongyu Wang,
    Jing Liu
}
\affiliations{
    Guangzhou Institute of Technology, Xidian University, Guangzhou, China\\
    {zeng\_zhi\_rong}@stu.xidian.edu.cn,
    xtliu@xidian.edu.cn,
    22171214833@stu.xidian.edu.cn,
    22171214782@stu.xidian.edu.cn,
    neouma@mail.xidian.edu.cn
    
}

\usepackage{bibentry}
\begin{document}

\maketitle

\begin{abstract}
%  Many outstanding RGB-T trackers have achieved excellent results by leveraging inter-modal fusion. However, these trackers either neglect temporal information or not fully utilizing it, failing to effectively balance multi-modal information and temporal information. To address this, we propose a novel Cross Fusion RGB-T Tracking architecture (CFBT) that allows the RGB and TIR modalities of both the initial template and the online template to fully participate in tracking. Additionally, we design three novel cross spatio-temporal information fusion modules: Cross Spatio-Temporal
% Augmentation Fusion (CSTAF), Cross Spatio-Temporal Complementarity Fusion (CSTCF), and Dual-Stream Spatio-Temporal Adapter (DSTA). CSTAF leverages a cross-attention mechanism to fully enhance the feature representation of the template, CSTCF utilizes complementary information between different branches to enhance target features and suppress background features, and DSTA adopts the adapter concept to adaptively fuse complementary information from multiple branches within the transformer layer, using the RGB modality as a medium. Through this ingenious cross-fusion of multiple perspectives, we add only a minimal number of trainable parameters (less than 0.3\% of the total model parameters), enabling efficient fusion of temporal information. Extensive experiments on three popular RGB-T tracking benchmarks demonstrate that our method achieves new state-of-the-art performance.
Many state-of-the-art RGB-T trackers have achieved remarkable results through modality fusion. However, these trackers often either overlook temporal information or fail to fully utilize it, resulting in an ineffective balance between multi-modal and temporal information. To address this issue, we propose a novel Cross Fusion RGB-T Tracking architecture (CFBT) that ensures the full participation of multiple modalities in tracking while dynamically fusing temporal information. The effectiveness of CFBT relies on three newly designed cross spatio-temporal information fusion modules: 
% Our model consists of three cross spatio-temporal information fusion modules: 
Cross Spatio-Temporal Augmentation Fusion (CSTAF), Cross Spatio-Temporal Complementarity Fusion (CSTCF), and Dual-Stream Spatio-Temporal Adapter (DSTA). CSTAF employs a cross-attention mechanism to enhance the feature representation of the template comprehensively. CSTCF utilizes complementary information between different branches to enhance target features and suppress background features. DSTA adopts the adapter concept to adaptively fuse complementary information from multiple branches within the transformer layer, using the RGB modality as a medium.
% Through these ingenious fusions of multiple perspectives, we add only a minimal number of learnable parameters (less than 0.3\% of the total model parameters), 
These ingenious fusions of multiple perspectives introduce only less than 0.3\% of the total modal parameters, but they indeed enable an efficient balance between multi-modal and temporal information. Extensive experiments on three popular RGB-T tracking benchmarks demonstrate that our method achieves new state-of-the-art performance.

\end{abstract}

\section{Introduction}
Object tracking is a fundamental task in computer vision that aims to locate the positions of target objects throughout video sequences. It has wide applications in surveillance, autonomous driving, robotic navigation, and human-computer interaction \cite{alldieck2016context, dai2021tirnet, chen2017rgb}. In recent years, RGB-based trackers have rapidly advanced, leading to many outstanding contributions \cite{ostrack, seqtrack, odtrack}. RGB-based trackers can achieve satisfactory tracking performance under good visible light conditions. However, they often fail to accurately capture the target in extreme lighting, back-
ground clutter, and adverse weather conditions. In contrast, thermal infrared (TIR) imaging captures the heat emitted by objects, allowing it to maintain good imaging performance in extreme lighting environments. Nonetheless, TIR images generally lack the clarity of RGB images. Therefore, how to effectively combine the advantages of both modalities for tracking is a topic worth exploring.
% These RGB-based trackers utilize visual features such as color, texture, and shape to achieve accurate and stable object tracking in many standard scenarios. However, RGB-based trackers still face significant challenges in certain complex scenarios, such as extreme lighting, low light, and adverse weather conditions, where the quality of RGB images significantly deteriorates, causing the tracker to fail to accurately capture the target \cite{zhang2020object}. This is mainly because RGB imaging relies on visible light, and the intensity and distribution of visible light are easily affected by environmental factors. In such cases, Thermal Infrared (TIR) imaging sensors demonstrate their unique advantages. It generates images by capturing the heat emitted by objects, thereby maintaining good imaging performance in dark or extremely lit environments. This enables effective object tracking in scenarios where high-quality RGB images cannot be obtained. Despite this, TIR images have their limitations as well. Due to the lower resolution of thermal infrared sensors, TIR images generally do not reach the clarity of RGB images. This means that relying solely on TIR images for object tracking may not provide enough detail to distinguish the target from the background in complex scenes. Therefore, how to fully combine the advantages of both modalities for tracking is a topic worth exploring.

\begin{figure}[t]
\centering
\includegraphics[width=1\linewidth]{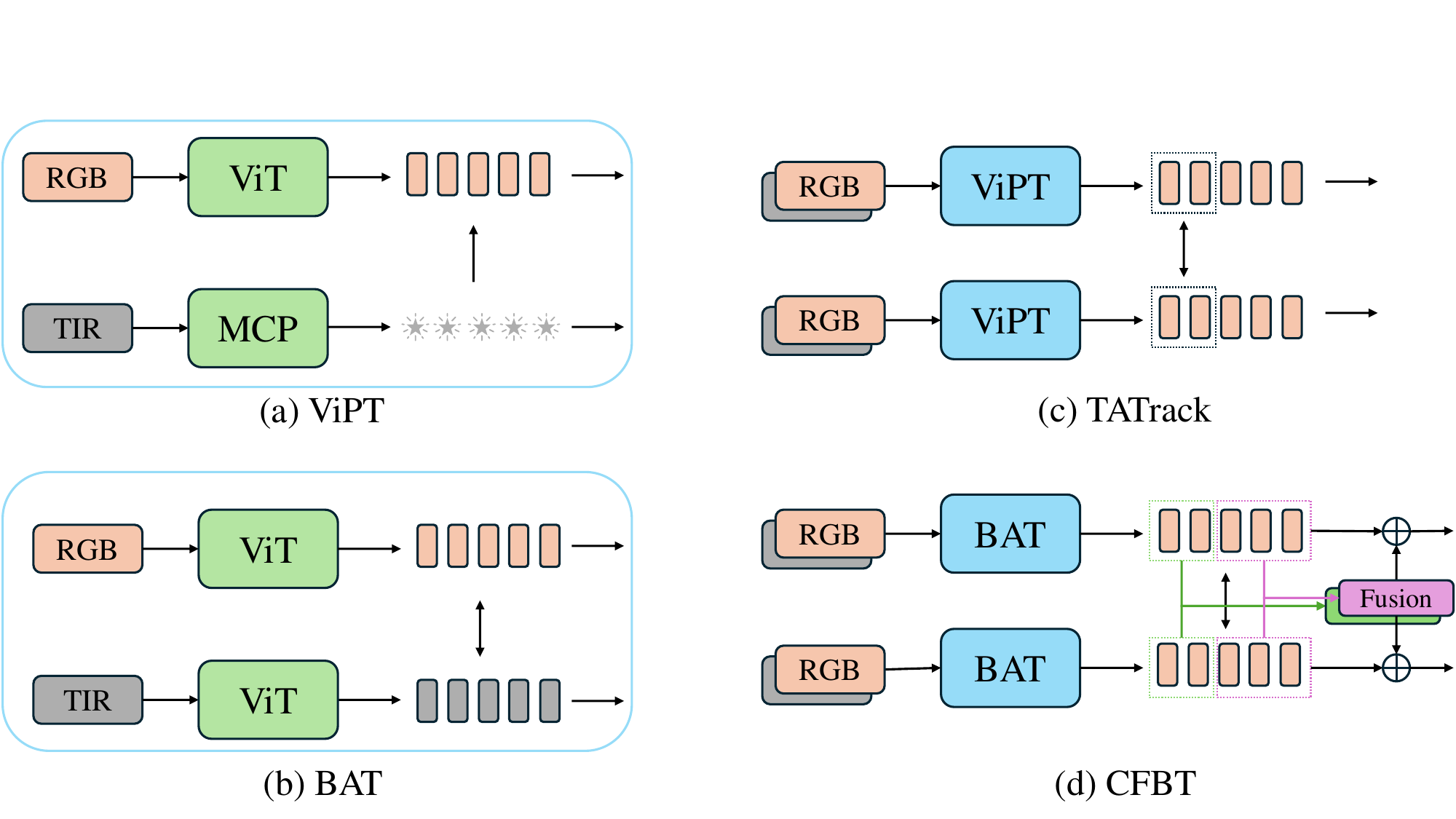} 
\caption{Differences between our RGB-T tracking approach
and previous ones. (a) Utilizing the TIR modality to assist the RGB modality. (b) Facilitating interaction between the TIR and RGB modalities. (c) Enhancing the templates of the RGB modality in different branches through interaction. (d) Using the RGB modality as a medium to enhance template interaction, complement the search regions through interaction, and transmit temporal information for deep cross-modal interaction.
}
\label{Different trackers}
\end{figure}

In previous RGB-T works, numerous excellent methods integrating RGB and TIR for tracking have emerged, such as ViPT \cite{ViPT}, TBSI \cite{TBSI}, and BAT \cite{BAT}. As shown in Figure 1(a, b), these RGB-T trackers focus solely on multi-modal information, neglecting temporal information during the tracking process. However, in addition to the challenges brought by complex scenes, targets often encounter occlusion and deformation, leading to significant discrepancies between the tracked target and the template, resulting in tracking failure. In such cases, relying solely on the initial template for tracking is insufficient to handle these scenarios. Utilizing temporal information for auxiliary tracking is equally important as multi-modal tracking. Unlike RGB-based tracking, which uses temporal information, RGB-T tracking needs to comprehensively consider multi-modal and temporal information. Employing a complex temporal information fusion module will add a significant number of additional parameters. Furthermore, the quantity of RGB-T tracking datasets is at least an order of magnitude smaller than that of RGB tracking datasets, making it difficult to achieve training results comparable to RGB-based tracking.
As depicted in Figure ~\ref{Different trackers}(c), current state-of-the-art trackers that utilize temporal information only focus on the interaction between templates in dual-branch framework, failing to fully exploit temporal information and introducing a substantial number of parameters. Neither approach effectively balances multi-modal and temporal information. Therefore, can we find a more effective method to achieve this?

Inspired by BAT, we find that an adapter architecture where both RGB and TIR modalities participate in tracking can better utilize multi-modal information. To this end, we propose a dual-stream adapter architecture that allows both the initial and online templates of the RGB and TIR modalities to participate in tracking, incorporating multi-modal fusion and temporal information fusion modules within the transformer layer. Additionally, we discover that solely relying on feature enhancement on the initial and online templates, as in TATrack, is insufficient. It not only depends on a complex temporal information fusion module but also neglects the feature representation of the search regions during tracking. Therefore, leveraging complementary information from multiple branches is more important. Unlike template enhancement methods, our model emphasizes utilizing this complementary information during tracking. We employ a cross-attention mechanism to dynamically enhance and complement effective information from multiple branches. Additionally, we adopt a lightweight hourglass structure that requires only a small number of additional learnable parameters to dynamically fuse information from these branches, making it easy to embed into the transformer layers. Experiments on the RGBT210 \cite{RGBT210}, RGBT234 \cite{RGBT234}, and LasHeR \cite{LasHeR} datasets validate the effectiveness of our CFBT framework. By training only a few parameters, CFBT achieves significant advantages compared to competitive methods.The main contributions are summarized as follows:
\begin{itemize}
\item We first propose a dual-stream adapter-based multi-modal tracking framework. It can dynamically perceive effective information in multiple branches for dynamic enhancement and complementarity. By adding only 0.259M learnable parameters, our model robustly handles multi-modal tracking in open scenarios.
\item We propose three novel cross spatio-temporal information fusion modules: CSTAF, CSTCF, and DSTA. CSTAF employs a cross-attention mechanism to fully enhance the feature representation of the templates. CSTCF leverages complementary information between different branches to strengthen the target features while suppressing the background features. DSTA adopts the concept of adapters, using the RGB modality as an intermediary within the transformer layer to adaptively fuse the complementary information from multiple branches.
\item Extensive experiments demonstrate that this method achieves state-of-the-art performance on three popular RGB-T tracking benchmarks.
\end{itemize}

\section{Related Works}
\subsection{Temporal Information Exploitation}
Object tracking is a continuous process that requires learning the appearance changes of the target from historical information. Relying solely on the initial template for tracking is challenging in the face of target deformation, occlusion, and other issues. With the development of object tracking, many excellent trackers have emerged in RGB-based tracking. EVPtrack \cite{EVPtrack} uses spatio-temporal markers to propagate information between consecutive frames, but it relies only on cues obtained from historical templates without fully utilizing temporal information. 
HIPTrack \cite{HIPtrack} proposes a historical prompt network, which provides prompts to the tracker using refined historical foreground masks and the target's historical visual features. However, it relies on complex feature extraction modules for additional feature extraction, making it difficult to apply in multi-modal tracking scenarios. ARTrack \cite{ARtrack} uses generative decoders for prediction, incorporating the target's historical trajectory, but these generative decoders introduce a large number of parameters. Additionally, ARTrack requires two-stage training, resulting in high training costs.
In multi-modal tracking scenarios, the complementarity between multiple modalities needs to be considered, which limits the use of temporal information. TATrack \cite{TATrack} introduces an online template for dual-branch tracking, leveraging temporal information and multi-modal complementarity to achieve significant improvements. However, it only uses the TIR modality to enhance the RGB modality without involving TIR in the actual tracking process. Furthermore, it relies on complex spatio-temporal interaction modules to fuse templates, which fails to consider the complementarity of temporal information across multiple branches. This approach not only does not fully utilize temporal information in the different branches but also introduces a substantial number of parameters.

\begin{figure*}[ht]
\centering
\includegraphics[width=1\linewidth]{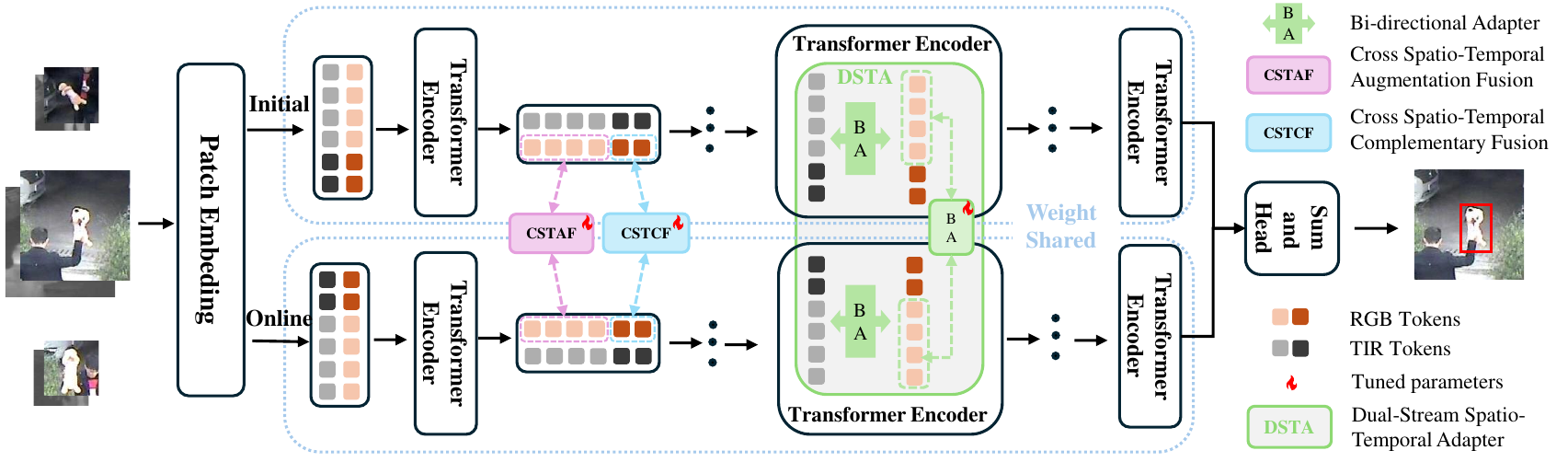} % e width.This setup will avoid overfull boxes.
\caption{The overall architecture of our proposed CFBT. First, we embed the image patches and then concatenate the initial template, online template, and search regions. These concatenated inputs are passed through $N$-layer of transformer encoders. Within these layers, the BA modules are inserted into each transformer encoder layer to enable cross-modal interaction. CSTAF and CSTCF modules are added at the 4th, 7th, and 10th layers to facilitate temporal information interaction. Additionally, DSTA modules are applied at the 5th, 6th, and 11th layers to further enhance deep temporal information interaction. Finally, the output features of the two branches are added and fed into the prediction head for final tracking result.
}
\label{fig:framework}
\end{figure*}

\subsection{Visual Prompt Learning in Multi-modal tracking}
Prompt Learning is an emerging technology in the field of natural language processing (NLP) that guides models to perform specific tasks by providing prompts. These prompts are typically learnable vectors pre-added to the input text to direct the model to generate specific outputs. This method significantly reduces the number of model parameters that need updating, thereby improving training efficiency and resource utilization. In previous multi-modal trackers, such as HMFT \cite{HMFT}, CAT \cite{CAT}, and APFNet \cite{APFNet}, parameters of RGB trackers are often loaded, followed by the design of new modules for full fine-tuning on multi-modal tracking datasets. However, multi-modal datasets are much smaller than RGB datasets due to the high costs of annotation, resulting in traditional fine-tuning-based trackers failing to achieve the desired results. 
ViPT \cite{ViPT} freezes the pre-trained base model and only trains a small number of parameters in the MCP module for modality interaction, achieving state-of-the-art performance while ensuring parameter efficiency. Subsequently, multi-modal tracking based on prompt learning is developing rapidly. BAT \cite{BAT} proposes an architecture based on adapters, enabling separate tracking of RGB and TIR modalities and achieving mutual cross-modal hints.
TBSI \cite{TBSI} proposes a new template-bridged search region interaction module, exploiting templates as a medium to bridge cross-modal interaction between RGB and TIR search regions by collecting and distributing target-related objects and environmental contexts. TATrack \cite{TATrack} introduces an online template update strategy, using spatio-temporal interaction to convey temporal information. 
However, previous works struggle to effectively balance multi-modal and temporal information, thus hindering robust tracking.
To overcome these limitations, we propose a novel Cross Fusion RGB-T Tracking architecture that ensures the full participation of multiple modalities in tracking while dynamically fusing temporal information from different perspectives.

\section{Methodology}
In this section, we will provide a detailed introduction to the proposed CFBT. Initially, we present an RGB-T tracking baseline that facilitates cross-modal interaction using Bi-directional Adapter. However, this baseline is limited in that it solely relies on match-based tracking with reference to the initial template without accommodating changes in the object's state. 
Subsequently, we extend this baseline to CFBT, which avoids extensive fine-tuning of the base model. Instead, it incorporates three lightweight cross spatio-temporal information fusion modules to effectively harness temporal and multi-modal data for precise target localization. The overall architecture of CFBT is illustrated in Figure ~\ref{fig:framework}.

\subsection{Tracking Baseline with Bi-directional Adapter}
The baseline is built on ViT \cite{ViT} and consists of three main components: ViT backbone, Bi-directional Adapter, and bounding box prediction head. Firstly, it crops RGB and TIR templates $I_{z}^{RGB}, I_{z}^{TIR} \in {R}^{3 \times W_z \times W_z} $ by given initial positions $B_{0}$ of the first frame. Subsequently, during tracking, it uses positions predicted from the previous frame for cropping and padding to obtain search regions $ I_{x}^{RGB}, I_{x}^{TIR} \in {R}^{3 \times W_x \times W_x} $. In the patch embedding layer of the transformer, these images are sliced into $ P \times P $ patches, flattened, and concatenated into 1D tokens. The initial template and the online template are respectively spliced with the search region to obtain the $ H^{RGB} \in {R}^{(N_z + N_x) \times D} $ and $ H^{TIR} \in {R}^{(N_z + N_x) \times D} $, where $ N_z $ and $N_x$ denote the number of tokens, and $D$ represents the token dimension.
The backbone network $(\mathcal{F})$ consists of $N$ layers of standard transformer blocks, responsible for feature extraction and interaction between template and search region. The prediction head $\mathcal{H}$ consists of a sequence of fully convolutional networks (FCN) designed to transform the extracted features into final bounding box predictions $B_{t}$. The traditional RGB-based object tracking can be represented as:
\begin{equation}
B_{t} = \mathcal{H}( \mathcal{F}(H_{t})).
\end{equation}
During the baseline tracking process, the aforementioned procedures will be independently applied to RGB and TIR images. Additionally, the Bi-directional Adapter (BA) will be integrated into both the multi-head self-attention and MLP stages of each transformer block to facilitate multi-modal interaction. The two processes are described below using the RGB modality as an example:
\begin{equation}
\begin{gathered}
H_{t}^{RGB} = H_{t}^{RGB} + \mathcal F^{attn}(H_{t}^{RGB}) +\mathcal F^{BA}(H_{t}^{TIR}), \\
H_{t}^{RGB} = H_{t}^{RGB} + \mathcal F^{MLP}(H_{t}^{RGB}) +\mathcal F^{BA}(H_{t}^{TIR}),
\end{gathered}
\end{equation}
where $\mathcal{F}^{attn}$ denotes the multi-head self-attention block, $\mathcal{F}^{MLP}$ refers to the multi-layer perceptron, and $\mathcal{F}^{BA}$ signifies the BA.

\begin{figure}[ht]
\centering
\includegraphics[width=0.6\linewidth]{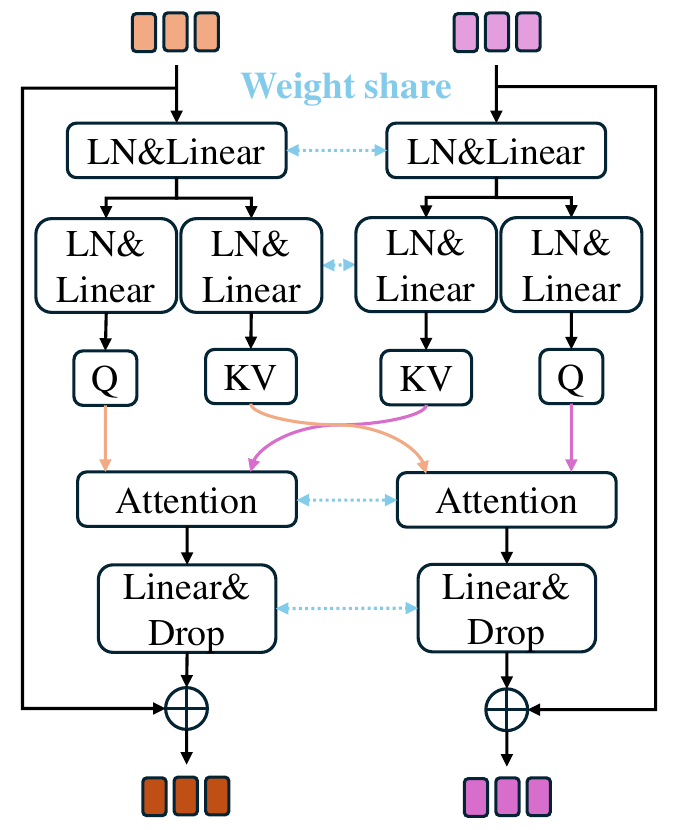}
\caption{
The overall framework of the Cross Spatio-Temporal Fusion module consists of three main components: down-projection linear layer, up-projection linear layer, cross attention layer. 
}
\label{SCTF}
\end{figure}

As shown in Figure~\ref{DSTA}, the BA consists of three linear layers. The input features are first down-projected using a linear layer, then passed through another linear projection layer, and finally up-projected back to the original dimensions. This simple architecture enables multi-modal interaction. If you want to know more about it, please refer to the BAT \cite{BAT}.

\subsection{Cross Fusion RGB-T Tracking}
The baseline method uses only the target in the first frame as a template, without accounting for changes in the target's appearance during tracking. This limitation makes it challenging to handle issues such as occlusion and deformation. Therefore, we extend the baseline by incorporating multi-modal and temporal information for tracking. In this section, we will provide a detailed explanation of how CFBT utilizes temporal information.

\textbf{Overall Architecture.}
Firstly, we introduce an online template strategy for dual-branch tracking. Specifically, we select reliable targets from the tracking results to create another template. Similar to the initial template, this new template also tracks the same search regions. In the two branches, the weights of the transformer encoder blocks and BA are shared. To facilitate temporal information interaction during dual-branch tracking, we incorporate CSTAF and CSTCF after the 4th, 7th, and 10th layers of the transformer encoder. CSTAF enhances the feature representation of both the template and the online template through spatio-temporal interaction, while CSTCF extracts complementary information from the search regions in the two branches, thereby enhancing target features and suppressing background features. Additionally, DSTA is integrated after the multi-head attention and MLP layers within the 5th, 6th, and 11th layers of the transformer encoder, further enabling adaptive feature fusion through the interaction between the search regions of different branches. It is noteworthy that the CSTAF in all layers share the same parameters and so does CSTCF.

\textbf{Cross Spatio-Temporal Fusion}
% Cross Spatio-Temporal Fusion module include CSTAF and CSTCF. 
The primary aim of CSTAF module is to integrate the features of the initial and online templates during tracking.
The cross-attention operation between the two not only enhances their feature representations, ensuring that both focus more on the target within the template, but also indirectly provides feature prompts from one to the other. It is noteworthy that we only fuse the two branches of the RGB modality. This is because the BA module already facilitates multi-modal interaction between the RGB and TIR modalities, allowing temporal information to be conveyed through the RGB modality alone. This approach not only conserves performance and space but also achieves the desired effect.

The CSTAF process is illustrated in Figure ~\ref{SCTF}. It begins by down-projecting the initial template $Z_i$ and the online template $Z_o$ to reduce their dimensions.
\begin{equation}
\begin{gathered}
Z_{i}^{origin}= Z_i\ ,\ Z_{o}^{origin}\ =\ Z_o,\\
Z_i=\ Down\left( Z_i \right)\ ,\ Z_o=Down\left( Z_o \right).
\end{gathered}
\end{equation}
This step is necessary because dual-branch fusion often contains redundant information, and we aim to minimize the excessive parameters introduced during interaction. Subsequently, the two templates undergo a cross-attention mechanism to interact and enhance their features.
\begin{equation}
Z_i\text{ , }Z_o=Cross\ Attention\left( Z_i, Z_o \right).
\end{equation}
After cross-attention, the templates are up-projected to restore their original dimensions.
\begin{equation}
Z_i=UP\left( Z_i \right)\ ,\ Z_o=UP\left( Z_o \right). 
\end{equation}
To prevent the model from learning unnecessary noise, we apply random dropout to the up-projected information before integrating it into the template via a residual connection.
% The overall process can be summarized as follows:
\begin{equation}
\begin{gathered}
Z_i=Z_{i}^{orign}+Drop\left( Z_i \right), \\ 
Z_o=Z_{o}^{orign}+Drop\left( Z_o \right), 
\end{gathered}
\end{equation}
where $Down(\cdot)$ denotes the down-projection linear layer, which reduces the dimensionality of the features by a factor of 16. Similarly, $UP(\cdot)$ represents the up-projection linear layer that restores the features to their original dimensions after the reduction.

The CSTCF module shares the same structure as CSTAF but serves different purposes. Since the initial template $Z_i$ and the online template $Z_o$ are images of the same object from different angles, they both interact with the same search regions. The fusion of these templates emphasizes correlation to enhance the feature representation of the target within the templates. This ensures that subsequent attention operations between the templates and the search regions focus more on the target.
\begin{figure}[ht]
\centering
\includegraphics[width=1\linewidth]{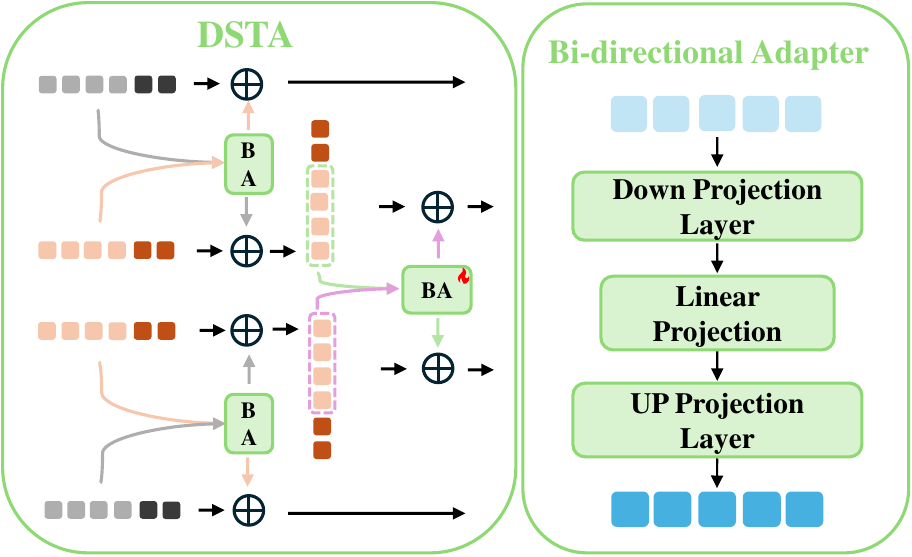}
\caption{The detailed architecture of DSTA and BA.
DSTA includes three BA modules, where we freeze the BA modules between modalities and only train the BA modules within the different branches. Each BA module consists of three linear layers: down-projection linear layer, linear projection layer, up-projection linear layer.
}
\label{DSTA}
\end{figure}
In contrast, the search regions $S_i$ and $S_o$ in the different branches come from the same image but interact with different templates. Due to their similar features, the fusion of these search regions emphasizes complementarity. Through the cross-attention operation, the search regions in each branch can better capture features that are not sufficiently represented in the other branch, thus progressively focusing more on the target.

\textbf{Dual-Stream Spatio-Temporal Adapter}
Inspired by BAT, we incorporate fusion modules within the transformer encoder layers to facilitate adaptive fusion, thereby enhancing the transfer of temporal information between the different branches. As depicted in Figure ~\ref{DSTA}, after interaction within each modality, we utilize the RGB modality as a conduit for the exchange of temporal information between the initial template branch and the online template branch. The following describes this process using the fusion after the multi-head attention block as an example:

First, the RGB modalities from the different branches separately enter a multi-head attention layer, and the outputs $H^{RGB}$ are interacted with $H^{TIR}$.
\begin{equation}
\begin{gathered}
H_{i}^{RGB} = H_{i}^{RGB} + \mathcal F^{attn}(H_{i}^{RGB}) +\mathcal F^{BA}(H_{i}^{TIR}),\\
H_{o}^{RGB} = H_{o}^{RGB} + \mathcal F^{attn}(H_{o}^{RGB}) +\mathcal F^{BA}(H_{o}^{TIR}).
\end{gathered}
\end{equation}
 Subsequently, $H^{RGB}$ is divided into template and search parts, resulting in $H^{t}$ and $H^{s}$, respectively. 
\begin{equation}
\begin{gathered}
H_{i}^{t} , H_{i}^{s} =split(H_{i}^{RGB}),\\
H_{o}^{t} , H_{o}^{s} =split(H_{o}^{RGB}).
\end{gathered}
\end{equation}
Here, $H^{t} $ remains unchanged, while $H_{i}^{s} $ undergoes adaptive fusion with $H_{o}^{s} $. Finally, the fused $H^{s} $ and $H^{t} $ are concatenated together.
\begin{equation}
\begin{gathered}
H_{i}^{RGB} =\ cat(H_{i}^{t},\ H_{i}^{s} + \mathcal F^{BA}(H_{o}^{s})),\\
H_{o}^{RGB} =\ cat(H_{o}^{t},\ H_{o}^{s} + \mathcal F^{BA}(H_{i}^{s})).
\end{gathered}
\end{equation}

\subsection{Objective Loss} 
The token sequence is initially transformed into a 2D spatial feature map via a series of fully convolutional networks (FCN). This transformation produces a target classification score map, indicating the location of the target, alongside the offset and normalized bounding box. The comprehensive loss function for CFBT is defined as:
\begin{equation}
L_{total} = L_{cls} + \lambda_1L_{iou} + \lambda_2L_1,
\end{equation}
where $L_{cls}$ represents the weighted focal loss for classification, while the generalized IoU loss $L_{iou}$ and L1 loss $L_1$ are employed for bounding box regression. The parameters $\lambda_1$ and $\lambda_2$ are trade-offs that balance these components.

\section{Experiments}

\subsection{Implementation Details}
We conduct experiments on three multi-modal tracking datasets: RGBT210, RGBT234, and LasHeR. The tracking performance is evaluated using five metrics: Precision Rate (PR), Maximum Precision Rate (MPR), Normalized Precision Rate (NPR), Success Rate (SR), and Maximum Success Rate (MSR). The CFBT framework is implemented in Python using the PyTorch library. Our model is trained on three NVIDIA RTX 3090 GPUs and one NVIDIA RTX A6000 GPU.

\textbf{Training.} We fine-tune our model on the LasHeR dataset, with a total batch size of 128. The fine-tuning process spans 25 epochs, with each epoch comprising 6 × $\rm{10^{4}}$ sample pairs. We utilize the AdamW optimizer \cite{AdamW} with a weight decay of $\rm{10^{-4}}$. The initial learning rate is set to 1 × $\rm{10^{-4}}$ and is reduced by a factor of 10 after 20 epochs. The search regions and templates are resized to 128 × 128 and 256 × 256, respectively. We initialize our model with the parameters from BAT, freezing all of these parameters, and train only the newly proposed modules.

\textbf{Inference.} We use the initial template, the online template, and the search region as inputs to CFBT. The online template is updated by default every 50 frames. The template with the highest target classification score within each interval is selected to replace the previous one.

\begin{table*}[t]
\centering
% \small 
\begin{tabular}{c|c|c|cc|cc|ccc}
\hline
\multirow{2}{*}{Method} & \multirow{2}{*}{Source} & \multirow{2}{*}{Temporal} & \multicolumn{2}{c|}{RGBT210 } & \multicolumn{2}{c|}{RGBT234 } & \multicolumn{3}{c}{LasHeR } \\ \cline{4-10} 
                        &            &             & MPR            & MSR           & MPR            & MSR           & PR      & NPR         & SR    \\ \hline
ProTrack \cite{ProTrack} & ACM MM2022 & False             & 79.5          & 59.9         & 78.6          & 58.7         & 53.8    & -           & 42.0   \\        
APFNet \cite{APFNet}     & AAAI2022 & False                & 79.9          & 54.9         & 82.7          & 57.9         & 50.0    & 43.9        & 36.2   \\ 
MFNet \cite{MFNet}       & IVC2022 & False                 & -             & -            & 84.4          & 60.1         & 59.7    & 55.4        & 46.7   \\ 
DMCNet \cite{DMCNet}     & TNNLS22  & False                & 79.7          & 55.9         & 83.9          & 59.3         & 49.0    & 43.1        & 35.5   \\ 
SiamMLAA \cite{SiamMLAA} & TMM2023  & False                & -             & -            & 78.6          & 58.4         & -       & -           & -      \\ 
LSAR \cite{LSAR}         & TCSVT2023& False                & -             & -            & 78.4          & 55.9         & -       & -           & -      \\ 
MACFT \cite{MACFT}       & Sensors2023 & False             & -             & -            & 85.7          & 62.2         & 65.3    & -           & 52.5   \\ 
RSFNet \cite{RSFNet}     & ISPL2023 & False                & -             & -            & 86.3          & 62.2         & 65.9    & -           & 52.6   \\ 
DFMTNet \cite{DFMTNet}   & IEEE Sens.J23 & False           & -             & -            & 86.2          & 63.6         & 65.1    & -           & 52.0   \\ 
CMD \cite{CMD}           & CVPR2023 & False                 & -             & -            & 82.4          & 58.4         & 59.0    & 54.6        & 46.4   \\ 
ViPT \cite{ViPT}         & CVPR2023 & False&83.5& 61.7& 83.5          & 61.7         & 65.1    & -           & 52.5   \\ 
TBSI \cite{TBSI}         & CVPR2023 & False                & 85.3 & {62.5} & 87.1 & 63.7 & {69.2}    & 65.7        & 55.6   \\ 
CAT++ \cite{CAT++}       & TIP2024 & False                 & 82.2          & 56.1         & 84.0          & 59.2         & 50.9    & 44.4        & 35.6   \\ 
AMNet \cite{AMNet}       & TCSVT2024 & False               & -             & -            & 85.5          & 60.7         & 61.4    & 55.9        & 47.7   \\ 
SDSTrack \cite{sdstrack}       & CVPR2024 & False               & -          & -         & 84.8          & 62.5         & 66.7    & -        & 53.1   \\ 
OneTracker \cite{onetracker}         & CVPR2024 & False               & -          & -         & 85.7          & 64.2         & 67.2    & -        & 53.8   \\ 
BAT \cite{BAT}           & AAAI2024 & False                & -             & -            & 86.8          & 64.1         & \textbf70.2    & 66.4          & 56.3   \\ 
TransAM \cite{AMNet}     & TCSVT2024 & False               & -             & -            & {87.7} & {65.5} & {70.2} & 66.0        & 55.9   \\ \hline
CMPP \cite{CMPP} & CVPR2020 & True             & -          & -         & 82.3          & 57.5         & -    & -           & -   \\
TATrack \cite{TATrack}   & AAAI2024  & True              & 85.3 & 61.8         & 87.2          & 64.4        & 70.2    & 66.7 & 56.1   \\ 
STMT \cite{STMT}         & Arxiv2024 & True    & 83.0          & 59.5         & 86.5          & 63.8         & 67.4    & 63.4        & 53.7   \\ \hline
CFBT                  & -  & True                      & \textbf{87.7} & \textbf{63.0} & \textbf{89.9} & \textbf{65.9} & \textbf{73.2} & \textbf{69.5} & \textbf{58.4} \\ \hline
\end{tabular}
\caption{Comparison with state-of-the-art trackers on RGBT210, RGBT234 and LasHeR testing set.}
% Higher values indicate better performance. The best three results are shown in \color{red}{red} \color{black}{, } \color{green}{green} \color{black}{and} {blue} \color{black}{fonts}.
\label{table_rgbt210_rgbt234_and_lasher_results}
\end{table*}

\subsection{Comparisons}
Our model is compared against 21 competing methods. Quantitative comparisons are reported in Table~\ref{table_rgbt210_rgbt234_and_lasher_results}, with qualitative evaluation results presented in Figure~\ref{Visualization}.

\subsubsection{Quantitative Evaluation on RGBT210.}
RGBT210 is a classic RGB-T tracking benchmark comprising 210 sequences, with a maximum of 8K frames per video pair.
As shown in Table~\ref{table_rgbt210_rgbt234_and_lasher_results}, the best-performing TATrack based on spatio-temporal tracking achieves comparable results to TBSI, which utilizes only multi-modal fusion in RGBT210. This indicates that relying solely on initial and online templates for temporal information propagation is insufficient. TATrack struggles to effectively harness temporal information, limiting its generalization capability. In contrast, CFBT utilizes temporal cues through template enhancement, complementary search regions, and temporal prompts in the transformer layer, outperforming previous trackers. It achieves a 2.5\% improvement over TATrack and TBSI on MPR, a 0.5\% improvement over TBSI on MSR, and a 1.2\% improvement over TATrack.

\subsubsection{Quantitative Evaluation on RGBT234.}
RGBT234 is an expanded dataset based on RGBT210, offering 234 sequences of aligned RGB and TIR videos. As shown in Table~\ref{table_rgbt210_rgbt234_and_lasher_results}, most leading trackers have been evaluated on RGBT234, with CFBT consistently maintaining a top position. With the increased scale of the dataset, CFBT's advantages become more pronounced, surpassing TATrack by 2.7\% on MPR and leading by 1.5\% on MSR. The success of RGB-T trackers that utilize temporal information underscores the importance of these temporal cues. How to fully leverage temporal information without introducing a significant number of additional parameters is a crucial topic worthy of discussion.

\subsubsection{Quantitative Evaluation on LasHeR.}
LasHeR is a large-scale, high-diversity benchmark designed for short-term RGB-T tracking. It comprises 1,224 pairs of visible and thermal TIR videos, totaling over 730,000 frame pairs. 
Compared to the previous two datasets, LasHeR features a larger scale and a greater variety of sequences, providing a better platform to showcase the model's practical capabilities. According to the Table~\ref{table_rgbt210_rgbt234_and_lasher_results}, CFBT achieves outstanding results across three metrics: PR, NPR, and SR. It leads by 3\% in PR, 3.2\% in NPR, and 2.1\% in SR over the second-ranked tracker. Moreover, CFBT introduces only 0.259M parameters, significantly fewer than TATrack's 22M, thereby fully demonstrating CFBT's superiority.

\begin{figure}[ht]
\centering
\includegraphics[width=1\linewidth]{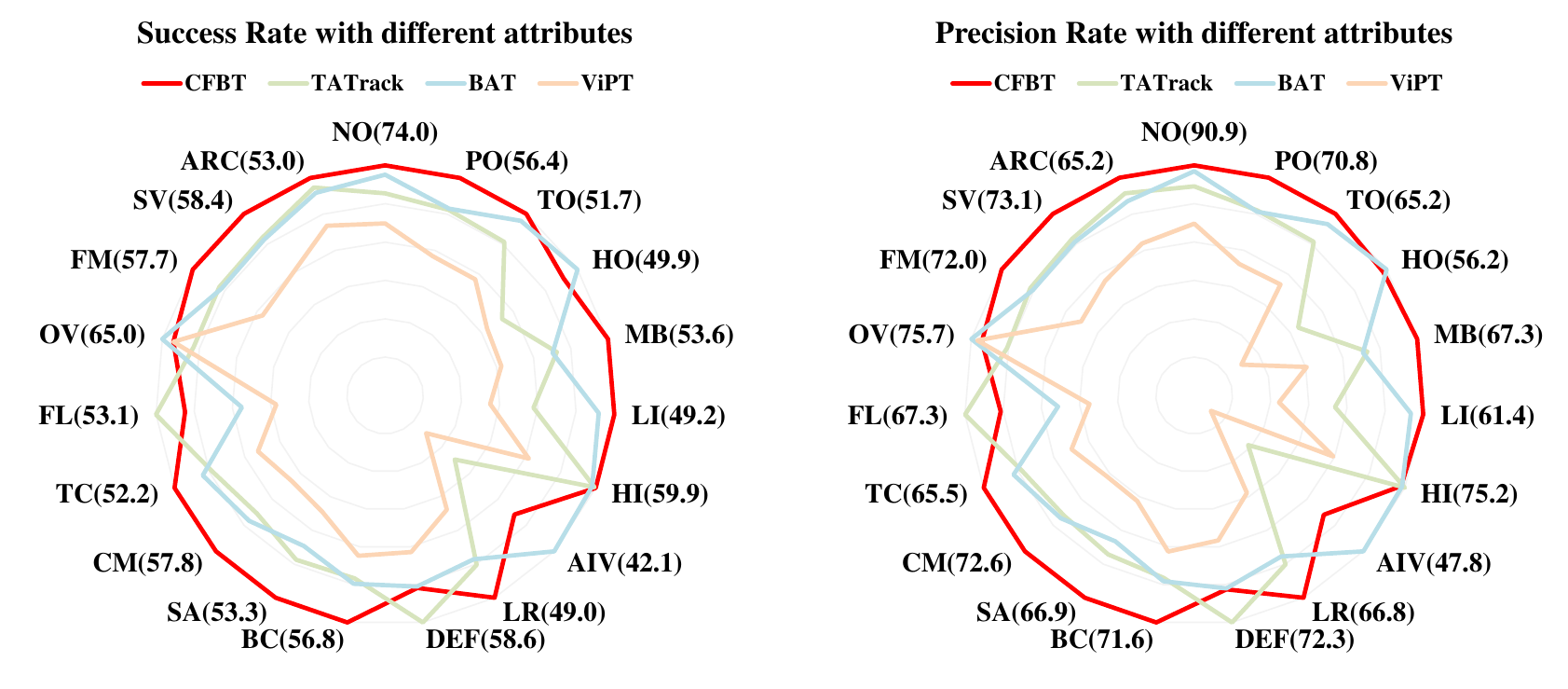} 
\caption{Further comparisons of CFBT and the competing methods under different attributes in the LasHeR dataset.}
\label{LasHeR_attributes}
\end{figure}

\subsubsection{Further Comparisons under Different Attributes.}
To further explore the comprehensive capabilities of the model, we conduct tests on 19 attributes using the LasHeR dataset and compare them with state-of-the-art methods. As shown in Fig~\ref{LasHeR_attributes}, it is evident that CFBT outperforms other advanced methods in the majority of attributes. Particularly noteworthy is its performance in challenges related to target appearance such as scale variation (SV), motion blur (MB), background clutter (BC), similar appearance (SA), fast motion (FM), and camera moving (CM), which further demonstrates CFBT's ability to utilize temporal information. In addition, compared to the similar type of TATrack, CFBT shows a 7.1\% improvement in low illumination (LI), a 7.5\% improvement in heavy occlusion (HO), and a 6.9\% improvement in abrupt illumination variation (AIV), further proving CFBT's superior generalization capability.

% \begin{figure*}[ht]
\begin{figure}[ht]
\centering
\includegraphics[width=1\linewidth]{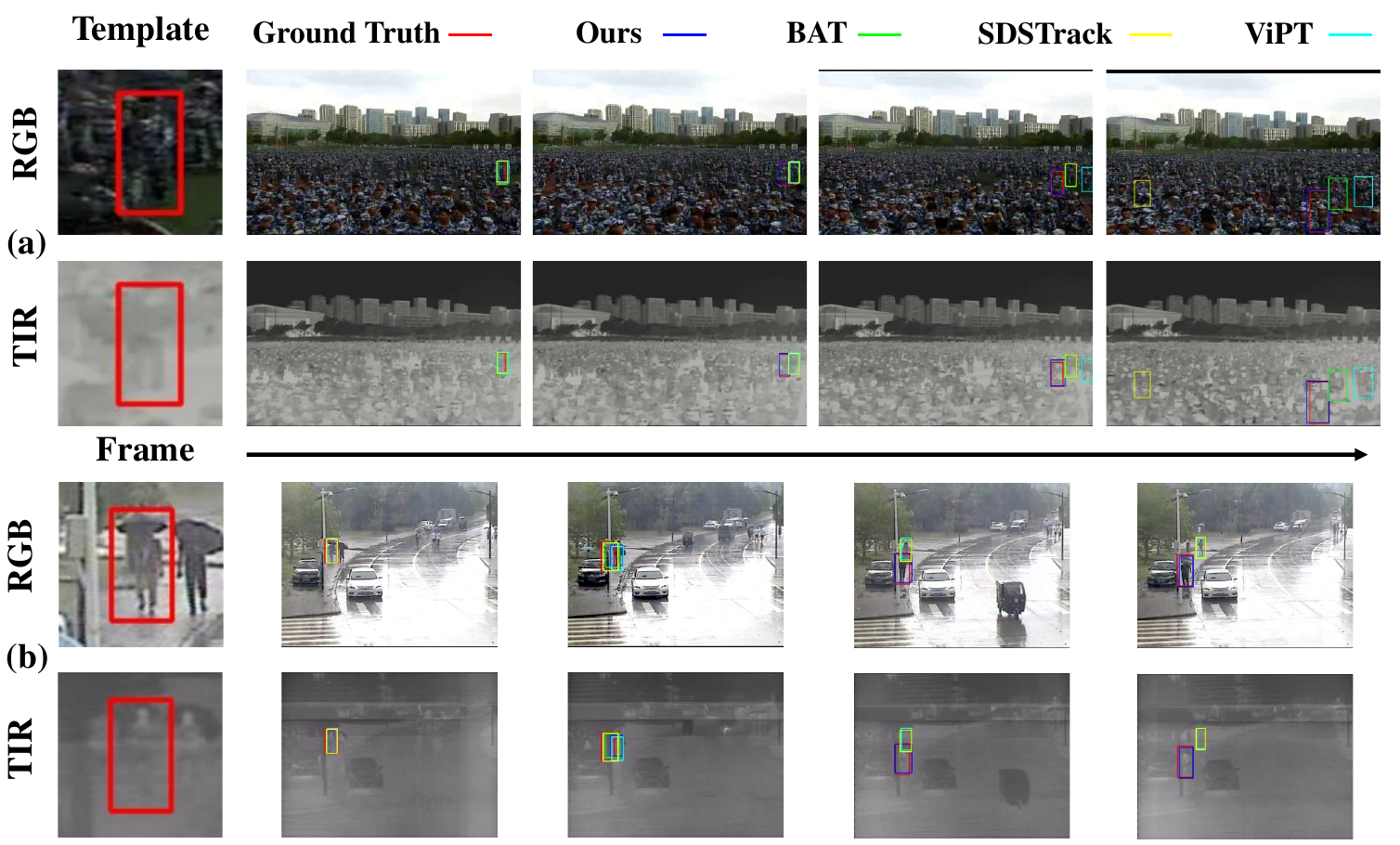}
\caption{Visualization of tracking results. The red rectangles indicate target objects. By incorporating temporal information, our method addresses challenges such as similar objects and occlusion, achieving the best performance.}
\label{Visualization}
\end{figure}
% \end{figure*}

\subsubsection{Qualitative Evaluation.}
Tracking with only the first frame as a template fails to capture sufficient appearance information of the target. In scenarios like Figure~\ref{Visualization}(a), where there are numerous similar objects, this lack of appearance information can easily lead to tracking errors. In contrast, CFBT fully leverages temporal information by continuously learning the appearance variations of the target and dynamically integrating effective information from different branches. This enables CFBT to maintain efficient tracking even in complex scenarios with similar objects.
Furthermore, in occlusion scenarios like Figure~\ref{Visualization}(b), where a large portion of the target is obscured, relying solely on the appearance information from the initial template makes it difficult to accurately determine the target's position from the limited visible regions. CFBT, with its exceptional cross spatio-temporal fusion, demonstrates robust performance under such challenging conditions.

\subsection{Ablation Study}
To verify the effectiveness of the main components, we conduct a detailed ablation study on the LasHeR dataset.

\begin{table}[t]
    \centering
    \begin{tabular}{lccc}
        \toprule
        Method        & PR & SR & Params$^{*}$ \\
        \midrule
        Baseline     & 70.2  & 56.3 & - \\
        +STI (\ding{172})          & 71.2     & 56.9 & 21.264M   \\
        +CSTAF (\ding{173})   & 71.7  & 57.2 & 0.090M \\
        +CSTCF (\ding{174})      & 71.6  & 57.3 & 0.090M \\
        +CSTAF\&CSTCF (\ding{175})  & 72.3 & 57.9 & 0.180M\\
        \midrule
        +CSTAF\&CSTCF\&DSTA  & \textbf{{73.2}} & \textbf{{58.4}} & 0.259M\\
        \bottomrule
    \end{tabular}
    \caption{Component analysis on LasHeR dataset.
    STI: Spatio-temporal Interaction of TATrack.
    Params$^{*}$: trainable parameters.
    }
    \label{component_analysis}
\end{table}

\subsubsection{Component Analysis.}As shown in Table~\ref{component_analysis}, To further validate the effectiveness of CFBT, we incorporate the STI module (\ding{172}) from TATrack into our baseline. This step ensures a fair comparison and eliminates potential errors introduced by differences in the baseline. CSTAF, enhanced through a hourglass structure and cross fusion, introduces only a small number of parameters yet achieves superior results. Specifically, CSTAF(\ding{173}) outperforms STI by 0.5\% in PR and performs comparably in SR.

Furthermore, to explore the contributions of each module to the overall structure, we individually trained and tested models using CSTAF (\ding{173}), CSTCF (\ding{174}), CSTAF + CSTCF (\ding{175}) and CSTAF + CSTCF + DSTA (CFBT). Compared to the baseline, \ding{173} increases performance by 1.5\% in PR and 0.6\% in SR. \ding{174} improves PR by 1.5\% and SR by 0.9\%. \ding{175} enhances PR by 2.1\% and SR by 1.6\%. Finally, CFBT results in a PR increase of 3.0\% and an SR increase of 2.1\%. These experimental results demonstrate that each module of CFBT has played a crucial role. Under equivalent conditions, CFBT achieves superior performance while using fewer parameters.
% \begin{table}[!t]
%     \centering
%     \setlength\tabcolsep{5pt}
%     \begin{tabular}{c|cccc|c|c}
%     \hline
%         Model & STI & CSTAF & CSTCF & DSTA & PR & SR\\ \hline
%         Baseline & ~ & ~ & ~ & ~ & 70.2 & 56.3 \\ 
%         \ding{172} & \checkmark & ~ & ~ & ~  & 71.2 & 56.9 \\ 
%         \ding{173} & ~ & \checkmark & ~ & ~&\textbf{\textcolor{blue}{71.7}} & 56.9  \\ 
%         \ding{174} & ~ & ~ & \checkmark & ~ &\textbf{\textcolor{blue}{71.7}} & \textbf{\textcolor{blue}{57.2}} \\
%         \ding{175} & ~ & \checkmark & \checkmark & ~ & \textbf{\textcolor{green}{72.3}} & \textbf{\textcolor{green}{57.9}} \\ \hline
%         CFBT & ~ & \checkmark & \checkmark & \checkmark & \textbf{{73.2}} & \textbf{{58.4}} \\ \hline
%     \end{tabular}
%     \caption{Component analysis on LasHeR dataset.\\
%     STI: Spatio-temporal Interaction of TATrack.}
%     \label{component_analysis}
% \end{table}

% \begin{table}[!ht]
%     \centering
%     \begin{tabular}{ccc|c|c}
%     \hline
%         \multicolumn{3}{c|}{Inserting Layers} & \multirow{2}{*}{Precision} & \multirow{2}{*}{Success} \\
%         4 & 7 & 10 & ~ & ~ \\ \hline
%         \checkmark & ~ & ~ & 72.4 & 57.8  \\ 
%         \checkmark & \checkmark & ~ & 71.1 & 56.8\\ 
%         \checkmark & ~ & \checkmark & 72.2 & 57.7\\ \hline
%         \checkmark & \checkmark & \checkmark & \textbf{73.2} &
%         \textbf{58.4}\\ \hline
%     \end{tabular}
%     \caption{Inserting layers of the proposed CSTAF and CSTCF.}
%     \label{Layers of CSTF}
% \end{table}
\begin{table}[!ht]
    \centering
    \begin{tabular}{ccc|ccc|c|c}
    \hline
        \multicolumn{3}{c|}{CSTF} &
        \multicolumn{3}{c|}{DSTA} & \multirow{2}{*}{Precision} & \multirow{2}{*}{Success} \\ \cline{1-6} 
        4 & 7 & 10 & 5 & 6 & 11 ~ & ~ \\ \hline
        \checkmark & ~ & ~ &
        \checkmark &\checkmark &\checkmark&
        72.4 & 57.8  \\ 
        \checkmark & \checkmark & ~ &
        \checkmark &\checkmark &\checkmark&
        71.1 & 56.8\\ 
        \checkmark & ~ & \checkmark &
        \checkmark &\checkmark &\checkmark&
        72.2 & 57.7\\ \hline
        \checkmark &\checkmark &\checkmark&
        \checkmark & ~ & ~ &
        71.5 & 57.2  \\ 
        \checkmark &\checkmark &\checkmark&
        \checkmark & \checkmark & ~ &
        73.1 & 58.3\\ 
        \checkmark &\checkmark &\checkmark&
        \checkmark & ~ & \checkmark &
        72.4 & 58.0\\ \hline
        \checkmark & \checkmark & \checkmark &
        \checkmark &\checkmark &\checkmark&
        \textbf{73.2} &\textbf{58.4}\\ \hline
    \end{tabular}
    \caption{Effects of inserting different layers.}
    \label{Layers of CSTF}
\end{table}

\subsubsection{Effects of inserting different layers.}
As shown in Table~\ref{Layers of CSTF}, we experimentally explore the effects of CSTF (CSTAF, CSTCF) and DSTA at various layers and summarized the results. The experiments demonstrate that although the modules share parameters across different layers, CSTF can still achieve effective temporal prompts by introducing only a small number of additional parameters. Moreover, the parameters in each layer of DSTA are independent. As the number of parameters increases, the tracking performance improves, but the rate of improvement slows down.

\section{Conclusion}
In this paper, we propose a novel cross-modality RGB-T tracking framework named CFBT. It avoids bias towards any specific modality or template and thoroughly exploits various types of information. Additionally, we design three lightweight fusion modules that integrate multi-modal and temporal information from the perspectives of feature representation of templates, complementary information from search regions, and temporal prompts in the transformer layer.
 Our approach achieves state-of-the-art performance on several mainstream RGB-T datasets by introducing only a minimal number of additional training parameters. Through this study, we highlight the untapped potential of temporal information in tracking applications, and we hope to stimulate further interest in this area. In future work, we aim to explore more efficient ways to harness temporal information for tracking across an broader range of modalities.

\section*{Acknowledgments}
This work is supported by National Defense Basic Scientific Research Project JCKY2021208B023, Guangzhou Key Research and Development Program 202206030003, and Guangdong High-level Innovation Research Institution Project 2021B0909050008.

\bibliography{aaai25}

\end{document}